\documentclass[10pt,twocolumn,letterpaper]{article}

\usepackage{wacv}              

\usepackage{graphicx}
\usepackage{amsmath}
\usepackage{amssymb}
\usepackage{booktabs}
\usepackage{dsfont}
\usepackage{multirow}
\usepackage{pifont}
\newcommand{\bftab}{\fontseries{b}\selectfont}
\usepackage{stackengine}
\usepackage{scalerel}
\newcommand\pluscross{\scalerel*{\stackinset{c}{}{c}{}{$+$}{$\times$}}{+}}

\usepackage{color, colortbl}

\usepackage[pagebackref,breaklinks,colorlinks]{hyperref}

\usepackage[capitalize]{cleveref}
\crefname{section}{Sec.}{Secs.}
\Crefname{section}{Section}{Sections}
\Crefname{table}{Table}{Tables}
\crefname{table}{Tab.}{Tabs.}

\def\wacvPaperID{518} 
\def\confName{WACV}
\def\confYear{2024}

\begin{document}

\title{SemST: Semantically Consistent Multi-Scale Image 
Translation via Structure-Texture Alignment}

\author{Ganning Zhao\\
University of Southern California\\
{\tt\small ganningz@usc.edu}
\and
Wenhui Cui\\
University of Southern California\\
\and
Suya You\\
DEVCOM Army Research Laboratory \\
\and
C.-C. Jay Kuo\\
University of Southern California
}
\maketitle


\begin{abstract}

Unsupervised image-to-image (I2I) translation learns cross-domain image
mapping that transfers input from the source domain to output in the
target domain while preserving its semantics.  One challenge is that
different semantic statistics in source and target domains result in
content discrepancy known as semantic distortion. To address this
problem, a novel I2I method that maintains semantic consistency in
translation is proposed and named SemST in this work. SemST reduces
semantic distortion by employing contrastive learning and aligning the
structural and textural properties of input and output by maximizing
their mutual information. Furthermore, a multi-scale approach is
introduced to enhance translation performance, thereby enabling the
applicability of SemST to domain adaptation in high-resolution images.
Experiments show that SemST effectively mitigates semantic distortion
and achieves state-of-the-art performance.  Also, the application of
SemST to domain adaptation (DA) is explored. It is demonstrated by
preliminary experiments that SemST can be utilized as a beneficial
pre-training for the semantic segmentation task. 

\end{abstract}

\section{Introduction}\label{sec:intro}

\begin{figure} [ht]
\begin{center} 
\includegraphics[height=5cm]{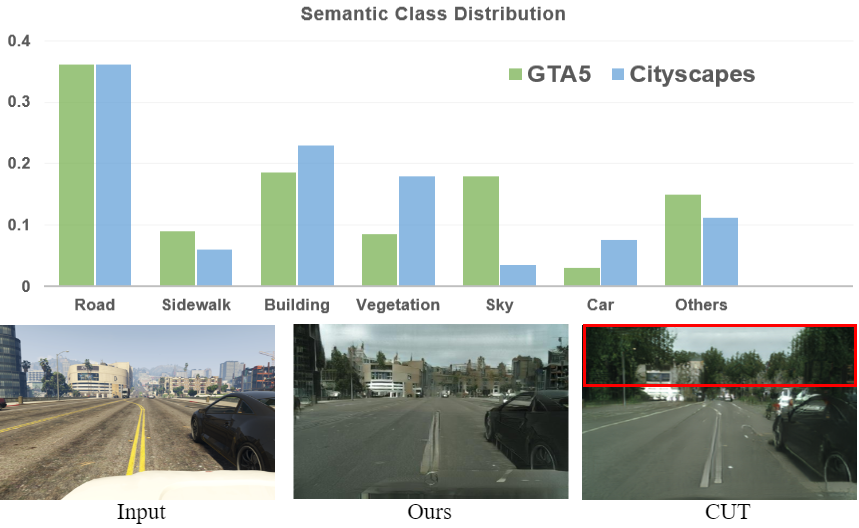}
\end{center}
\caption[1]{(top): The discrepancy in semantics distributions between
GTA5 and Cityscapes. More pixels of sky in GTA5, while more building,
vegetation, and car in Cityscapes. This significant difference in class
distributions introduces semantic distortion. (bottom): Illustration of
semantic distortion. In the GTA5-to-Cityscapes translation task, the sky
region is mistakenly transformed to vegetation by CUT
\cite{park2020contrastive}, due to more vegetation and less sky in
Cityscapes.  In contrast, our proposed SemST method preserves semantic
consistency between input and output.}\label{fig:dist}
\end{figure}

The objective of image-to-image (I2I) translation involves learning a
mapping from a source domain to a target domain. Specifically, it aims
at transforming images of the source style to those of the target style
with content consistency. While there is a domain gap, it can be
mitigated by aligning the distributions of the source and the target
domains. Nevertheless, disparities between class distributions of the
source and target domains result in semantic distortion (see Figure \ref
{fig:dist}); namely, different semantics of correspondent regions
between input and output.  The semantic distortion could potentially
impact the efficacy of downstream tasks, such as semantic segmentation
or object classification. 

Early works \cite{shrivastava2017learning, atapattu2019improving}
employed adversarial training to align distributions in different
domains with limited success. Since then, various techniques have been
developed to accomplish this task.  Bidirectional structures that
ensured cycle consistency were proposed in
\cite{zhu2017unpaired, yi2017dualgan, kim2017learning}. However, their
strict constraint of bijective projection could result in distortion.
Although one-sided image translation \cite{benaim2017one,
fu2019geometry, amodio2019travelgan} offers an alternative, its
semantic distortion remains to be a significant problem. Recently,
contrastive-learning-based methods were proposed, e.g.,
\cite{park2020contrastive}. Despite a large amount of effort, such as
leveraging more powerful loss functions \cite{yeh2022decoupled,
chen2021simpler}, mining informative positive/negative samples
\cite{wang2021instance, hu2021adco, robinson2020contrastive}, and
integrating various methods \cite{zhao2023unsupervised}, the capability
of the proposed methods in refining synthetic images and/or domain
adaptation remains limited and semantic distortion still exists. 

Besides image translation, reducing semantic distortion finds
applications in unsupervised domain adaptation (UDA). Training deep
neural networks (DNNs) in semantic segmentation demands expensive and
labor-intensive data labeling. It is desired to train DNNs on source
datasets containing existing (or more affordable) annotations and deploy
them on unlabeled target datasets. The main challenge in UDA is domain
shift, the discrepancy between the source and target domains. Extensive
efforts have been exerted to resolve this issue by aligning features
between the two domains. Since the domain gap in the image space
limits performance, researchers have recently turned to translating
images between domains and then aligning features from images.  This new
direction is proven advantageous \cite{li2019bidirectional,
musto2020semantically, ma2021coarse, ma2022i2f, hoffman2018cycada}.
However, current image translation approaches are usually applied to
images of low-resolution or downsampled to low-resolution, which
inevitably restricts performance in UDA that require high-resolution
images. Recent work \cite{hoyer2022hrda} demonstrates the performance
degradation when training UDA on images downsampled to low resolution. 

In this work, we propose a novel contrastive-learning-based method that
alleviates semantic distortion by ensuring semantic consistency between
input and output images. This is achieved by enhancing inter-dependence
of structure and texture features between input and output by maximizing
their mutual information. In addition, we exploit multi-scale
predictions to boost the I2I translation performance by employing global
context and local detail information jointly to predict translated
images of superior quality, especially for high-resolution images. Hard
negative sampling is also applied to reduce semantic distortion by
sampling informative negative samples. For brevity, we refer to our
method as SemST. Experiments conducted on I2I translation across various
datasets demonstrate the state-of-the-art performance of the SemST
method. Additionally, utilizing refined synthetic images in different
UDA tasks confirms its potential for enhancing the performance of UDA. 

\section{Related Work}\label{sec:ref}

\subsection{Unsupervised Image-to-Image Translation} \label{sec:ref1}

Initial investigations of unsupervised adversarial learning have focused
on augmenting the realism of synthetic images while maintaining
annotation information \cite{shrivastava2017learning,
atapattu2019improving}.  They were primarily applied to simple grayscale
images such as eyes and hands. Nonetheless, these methods found limited
success when applied to more complex datasets. 

Extensive efforts have been made to preserve semantics between input and
output images.  The cycle consistency loss was employed in
\cite{zhu2017unpaired, yi2017dualgan, kim2017learning,
theiss2022unpaired}, which assumed a bijective translation function
between the source and target domains.  They enforced consistency
between an input image in the source domain and the reconstructed image,
inversely translated from the corresponding target domain image.
However, these methods require an additional generator/discriminator
pair. Besides, the bijective assumption could introduce distortions
\cite{park2020contrastive, lee2018diverse, wang2021instance,
theiss2022unpaired}. 

Alternatively, one-sided image translation methods have emerged. They
enforced geometry consistency between a source image and its transformed
counterpart in the target domain \cite{fu2019geometry} or ensured a
strong correlation between matched pairwise distances in individual
domains \cite{benaim2017one, zhang2019harmonic}. Furthermore, some
research aimed to reduce semantic distortion caused by mismatched
semantic statistics by imposing structure consistency
\cite{guo2022alleviating} or semantically robust loss
\cite{jia2021semantically}. However, many challenges still exist, including but not limited to semantic distortion, training instability, and limited applicability to high-resolution images.

\subsection{Contrastive Learning} 

Contrastive Learning (CL) has been applied to image translation,
offering a means to learn useful representations by exploring
relationships among positive and negative pairs
\cite{park2020contrastive}. One idea is to develop suitable loss
functions.  The InfoNCE loss \cite{oord2018representation} linked
corresponding patches and disassociates others through cross-entropy
loss, gaining popularity in a few follow-ups, say,
\cite{bachman2019learning, park2020contrastive, chen2020simple,
he2020momentum}. Improved loss functions were proposed to address issues
arising from small batch sizes \cite{chen2021simpler} and alleviate the
negative-positive coupling (NPC) effect \cite{yeh2022decoupled}. 

Another line of research focuses on hard negative mining. One can employ
techniques like using a negative sample generator
\cite{wang2021instance}, sampling negatives via the von Mises Fisher
distribution \cite{robinson2020contrastive}, or resorting to adversarial
training \cite{hu2021adco}. 

In addition, some studies aim to mitigate semantic distortion by
exploring semantic relations among samples. For instance, one can ensure
cross-domain consistency between positive and negative samples in source
and target domains \cite{jung2022exploring, wei2020co2,
zhao2023unsupervised}. The idea to encode hierarchical semantic
structures in the embedding space using the EM algorithm was tried and
reported in \cite{li2020prototypical}. 

\subsection{Unsupervised Domain Adaptation} 

Most work on unsupervised domain adaptation (UDA) has concentrated on
feature-level adaptation through adversarial models. Research on
image-level translation has received less attention. Recently, it has
been shown in \cite{li2019bidirectional, musto2020semantically,
shen2023study, hoffman2018cycada} that models trained on synthetic
images translated from real image domains can enhance performance
significantly in the semantic segmentation task.  This indicates that,
compared with feature-level alignment, the domain gap can be further
reduced by image-level alignment. It was also reported in
\cite{ma2021coarse, ma2022i2f} that both image-level and feature-level
alignments contribute to performance improvement of domain adaptation. 

\begin{figure*} [ht]
\begin{center} 
\includegraphics[height=8cm]{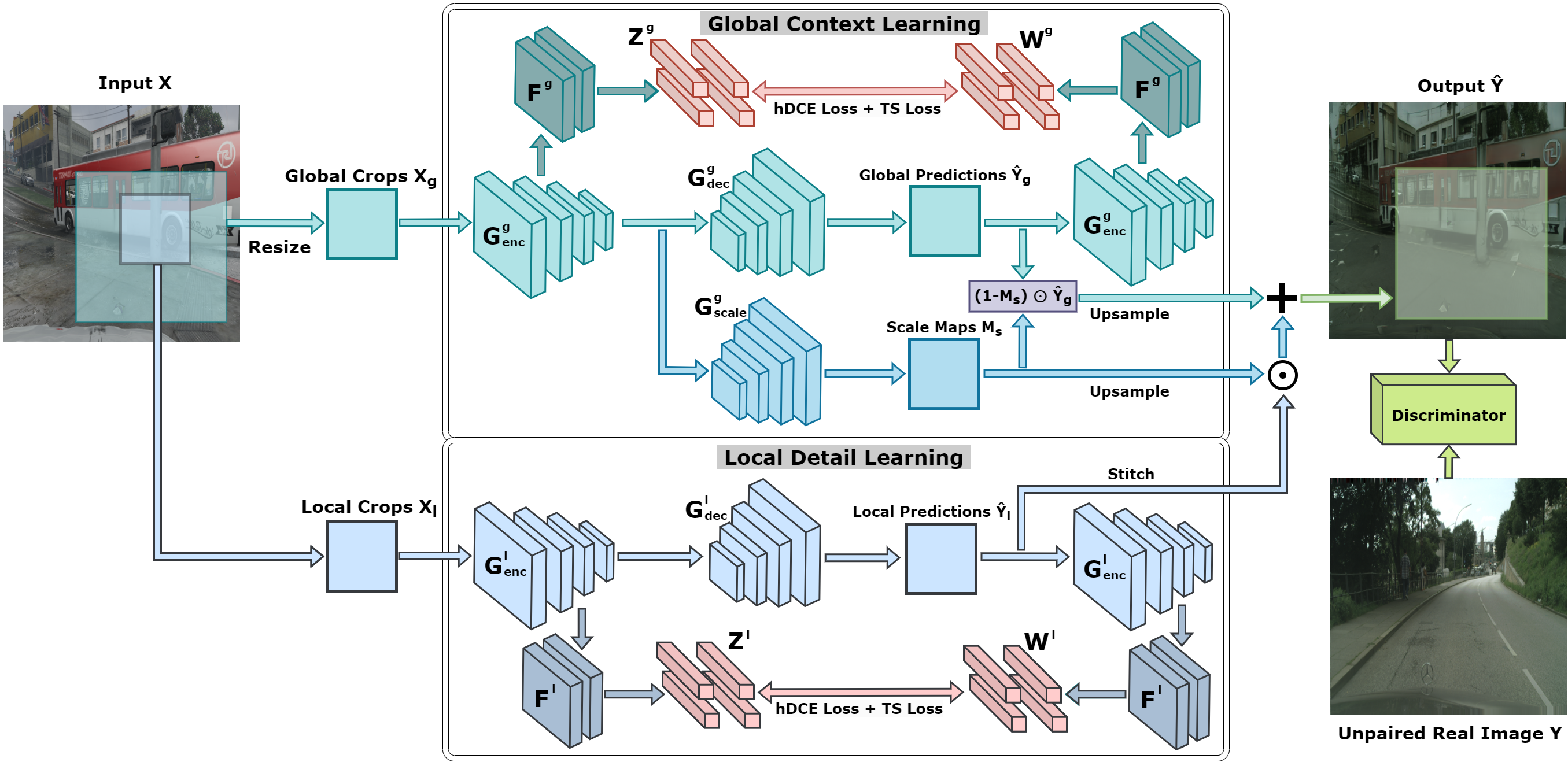}
\end{center}
\caption[1]{Overview of the proposed SemST method. It consists of a
global context and a local detail learning branch.  Global and local
crops, extracted from input images, are fed separately to global or
local learning branch. Generators of the encoder-decoder structure,
$G^g_{enc-dec}$ and $G^l_{enc-dec}$, learn global and local predictions,
respectively, which are finally fused together with a scale map $M_s$
that assesses the trustworthiness of global or local predictions.  In
each branch, embeddings of source and target domains are learned via the
shared fully connected layers {F} applied to encoders of input and
output, respectively. The hDCE and the TS losses are employed to align
semantics within the embeddings in both global and local branches. A
discriminator is trained to minimize the domain
gap.}\label{fig:pipeline}
\end{figure*} 

\section{Proposed SemST Method}\label{sec:contri}

\subsection{Motivation and System Overview}

The distributions of semantic labels are usually different in source and
target domains \cite{wei2020co2, yang2022mutual, jung2022exploring}, as
observed in Figure \ref{fig:dist} and prior methods
\cite{jia2021semantically, guo2022alleviating}, which not only leads to
pixels with error semantics but also adversely influences downstream
tasks that involve domain adaptation in the pipeline. In practical
applications, image semantics are correlated with low-level texture and
structure properties. For instance, sky, buildings and vegetation should
exhibit similar visual appearance in input and output domains. Thus, we
employ the joint texture (i.e., smooth or edge regions) and structure
information to maintain semantic consistency between input and output. 

The block diagram of the proposed SemST is depicted in Figure
\ref{fig:pipeline}. We will elaborate on the three components: 1)
structure and texture alignment for semantic consistency as indicated in
pink; 2) multi-scale prediction as indicated in gray; 3) semantics-aided
hard negative sampling. 

\subsection{Structure and Texture Alignment} 

\begin{figure} [ht]
\begin{center} 
\includegraphics[height=5cm]{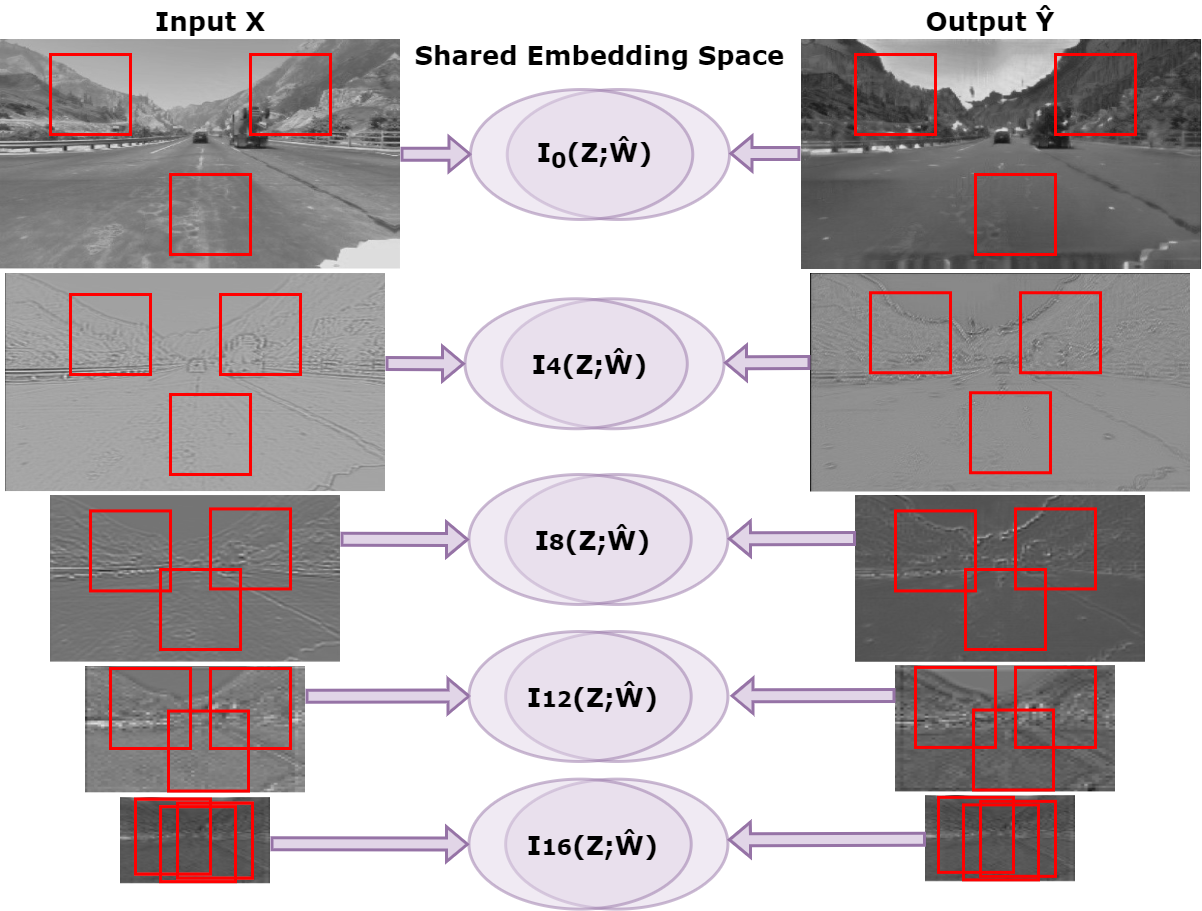}
\end{center}
\caption[1]{The mutual information, $I_x$ (where $x$ indicates layer
indexes), between the input and output embedding spaces is maximized to
maintain semantic consistency.}\label{fig:tsloss} \end{figure}

To alleviate semantic distortion, we propose a loss function to preserve
texture and structure consistency between input and output by maximizing
their mutual information. Generally speaking, embeddings in shallow
layers of higher resolutions capture the specific texture while embeddings in
deeper layers of lower resolutions reflect the generalized structure information as
illustrated in Fig. \ref{fig:tsloss}. The figure depicts embeddings from
shallow to deep layers obtained based on receptive fields of varying
sizes, encompassing the small-scale texture information to the
large-scale structure information. 

Embeddings in different layers of input and output images are extracted
and aligned for semantic consistency. To maintain their consistency, we
use mutual information to measure non-linear dependence between
input and output embeddings. For efficient backpropagation and robust
learning, we adopt the relative Squared-loss Mutual Information (rSMI). 

Mathematically, for the embedding, $z_i$, of the input image and the
embedding, $\hat{w}_i$, of the output image, we use $Z_i$ and $\widehat{W}_i$ to denote
their respective random variables. The texture-structure consistency (TS) loss is written as 
\begin{equation}\label{eq:tsloss}
L_{TS} = - \frac{1}{N} \sum_{i=1}^{N} \mbox{rSMI} (Z_i, \widehat{W}_i),
\end{equation}
where $N$ is the sample number. $\mbox{rSMI} (Z_i, \widehat{W}_i)$ is
computed by the relative Pearson (rPE) divergence
\cite{yamada2013relative, guo2022alleviating}, defined as
\begin{equation}
\mbox{rSMI} (Z_i, \widehat{W}_i) = D_{rPE}(P_{Z_i} \otimes P_{\widehat{W}_i} 
|| (P_{(Z_i, \widehat{W}_i)}),
\end{equation}
In practice, $\mbox{rSMI} (Z_i, \widehat{W}_i)$, is estimated by a
linear combination of kernel functions. It is solved by least-squares
density-difference estimation. As a result, the mutual information
estimator is in form of
\begin{equation}
\widehat{\mbox{rSMI}} (Z_i, \widehat{W}_i) = 2\hat{\alpha}^T\hat{h}-\hat{\alpha}^T
\hat{H}\hat{\alpha}-1.
\end{equation}
where $\hat{\alpha}$, $\hat{h}$, and $\hat{H}$ are parameters computed via 
least-squares density-difference estimation \cite{guo2022alleviating}. 

\subsection{Multi-Scale Framework}

Most prior art on I2I translation directly manipulated images downsampled to a
lower resolution, say, $256 \times 256$. However, this process
inevitably limits performance due to information loss in the
downsampling and subsequent upsampling of images back to their original
resolution. Besides, these methods failed to predict smaller objects
(e.g., poles and bikes) accurately and object borders with high quality.
The performance could be even more compromised when dealing with
intricate, high-resolution images containing objects of various scales.
Although training random crops could be a solution, it fails to learn
scene layout and relationships among objects, thereby introducing
errors.  To address these challenges, we propose a multi-scale framework
that concurrently predicts local crops on a small scale and global crops
on a larger scale. The model can learn detailed information with local
crops, e.g., small objects and intricate borders, and the
context information with global crops, e.g., layout and relationships
among objects. 

The above idea can be formalized as follows. We randomly crop the large global
crops, $X_g$, from input images and downsample ($T$) them to size $h_g \times w_g$:
\begin{equation}
X_g = T(X_{ori}[h_{g0}: h_{g1}, w_{g2}: w_{g3}]; h_g, w_g).
\end{equation}
Furthermore, small local crops $X_l$ of size $h_l \times w_l$ are 
randomly cropped from $X_{ori}$:
\begin{equation}
X_l = T(X_{ori}[h_{l0}: h_{l1}, w_{l2}: w_{l3}]; h_l, w_l).
\end{equation}
Global and local crops are predicted by the generator of the
encoder-decoder network, indicated by $\hat{Y}_g = G^g_{enc-dec}(X_g)$
and $\hat{Y}_l = G^l_{enc-dec}(X_l)$, respectively. Different generators
are employed for local and global crops, given the different scales of
their content and their requirement for distinct embedding spaces.
Notably, this approach allows flexibility in the sizes of local and global
crops, which can be equal or different. The overlapping predictions are
averaged to increase robustness when stitching images for the subsequent
fusion of local and global predictions. 

To integrate predictions across different scales effectively, scale
attention \cite{chen2017deeplab, chen2016attention, hoyer2022hrda} is
used to generate scale maps, denoted as $M_{s}$.  The scale maps assist
in determining which regions of the output should rely more on local or
global predictions. For instance, smaller objects and complex structures
such as trees and distant objects tend to rely on local predictions. In
contrast, simpler regions, such as roads and proximate buildings depend
more on global predictions. The final predictions are obtained by the
fusion of global and local crops from different scales in the form of
\begin{equation}
\hat{Y} = M_{s} \odot \hat{Y}_l + (1-M_{s}) \odot \hat{Y}_g.
\end{equation}


\subsection{Semantics-aided Hard Negative Sampling}

Easy negative samples are uncorrelated with the query
sample, diminishing the learning rate from more informative hard negative samples
\cite{yeh2022decoupled}. This cause the negative-positive coupling
(NPC) effect. The decoupled InfoNCE (DCE) loss is crucial in alleviating the NPC effect. 

Here, we adopt the DCE loss by excluding the positive pair from the
denominator of InfoNCE.  Concurrently, we sample hard negative samples
that exhibit semantic correlations with the query sample by the von
Mises-Fisher distribution \cite{robinson2020contrastive,
jung2022exploring}.  This approach ensures that negative samples and
query samples correspond to distinct latent classes, while also
maintaining a substantial semantic similarity, quantified through the
inner product. As a result, we can express this relationship as
\begin{equation}
z^- \sim q_\beta (z^-), \quad \textrm{where} \quad q_\beta (z^-) 
\varpropto e^{\beta z^T z^-} \cdot p(z^-),
\end{equation}
where $\beta$ is a concentration parameter that controls the
similarity of hard negative samples with query samples. Combining it
with the DCE loss, we obtain the hard Decoupled Contrastive Entropy (hDCE) loss:
\begin{equation}\label{eq:hDCE_loss}
L_{hDCE} = \mathbb{E}_{(z,\hat{w})} \left[-log \frac{\exp(\hat{w}^Tz/\tau)}
{N\mathbb{E}_{z^- \sim q_\beta}[\exp(\hat{w}^Tz^-/\tau)]} \right], 
\end{equation}
where $N$ is the number of negative patches and $\tau$ is a
temperature parameter that controls the strength of penalties on hard
negative samples.  Then, the approximate expectation can be obtained by 
\cite{robinson2020contrastive}
\begin{equation}
\begin{aligned}
& \mathbb{E}_{z^- \sim q_\beta}[\exp(\hat{w}^Tz^-/\tau)] \\ 
& = \frac{1}{N} \mathbb{E}_{z^- \sim p}
[\exp(\hat{w}^Tz^-/\tau) \exp(\beta z^Tz^-)].
\end{aligned}
\end{equation}

For the implementation, we reweight the negative samples by their correlations with the positive sample, $z^Tz^-$.

\begin{table*}[htbp]
\caption{Quantitative evaluations of our method and benchmarking
methods. The methods with $+$ are reproduced by
\cite{guo2022alleviating} and $\protect \pluscross$ are reproduced by us on a single GPU using the codes provided by the authors. The
best results are highlighted in bold.}
\label{tab:img_trans}
\centering
\scalebox{0.84} {
\begin{tabular}{c||ccc||ccc||ccc}\hline
\multirow{2}{*}{Methods} & \multicolumn{3}{c||}{GTA5 $\rightarrow$ Cityscapes} & \multicolumn{3}{c||}{Cityscapes Parsing $\rightarrow$ Image} & \multicolumn{3}{c}{Photo $\rightarrow$ Map} \\ \cline{2-10}
& pixel acc $\uparrow$ & class acc $\uparrow$ & mean IoU $\uparrow$ & 
pixel acc $\uparrow$ & class acc $\uparrow$ & mean IoU $\uparrow$ & RMSE $\downarrow$ & acc\%($\delta_1$) $\uparrow$ & acc\%($\delta_2$) $\uparrow$ \\ \hline
DRIT++ \cite{lee2020drit++} & 0.423 & 0.138 & 0.071 & \textbackslash & \textbackslash & \textbackslash & 32.12 & 29.8 & 52.1 \\ \hline
CycleGAN \cite{zhu2017unpaired} & 0.232$^{+}$ & 0.127$^{+}$ & 0.043$^{+}$ & 0.520 & 0.170 & 0.110 & \bftab 26.81 & 43.1 & \bftab 65.6 \\ \hline
GcGAN \cite{fu2019geometry} & 0.405$^{+}$ & 0.139$^{+}$ & 0.068$^{+}$ & 0.551 & 0.197 & 0.129 & 27.98 & 42.8 & 64.6 \\ \hline
CUT \cite{park2020contrastive} & 0.546$^{+}$ & 0.165$^{+}$ & 0.095$^{+}$ & 0.695$^{+}$ & 0.259$^{+}$ & 0.178$^{+}$ & 28.48$^{+}$ & 40.1$^{+}$ & 61.2$^{+}$ \\ \hline
SRUNIT \cite{jia2021semantically} & 0.581$^{\pluscross}$ & 0.135$^{\pluscross}$ & 0.079$^{\pluscross}$ & 0.505$^{\pluscross}$ & 0.175$^{\pluscross}$ & 0.096$^{\pluscross}$ &  28.40$^{\pluscross}$ & 41.2$^{\pluscross}$ & 60.5$^{\pluscross}$ \\ \hline
SRC \cite{jung2022exploring} & 0.597$^{\pluscross}$ & 0.187$^{\pluscross}$ & 0.111$^{\pluscross}$ & 0.787$^{\pluscross}$ & 0.259$^{\pluscross}$ & 0.207$^{\pluscross}$ & 27.98$^{\pluscross}$ & 41.2$^{\pluscross}$ & 61.7$^{\pluscross}$ \\ \hline
VSAIT \cite{theiss2022unpaired} & 0.603$^{\pluscross}$ & 0.179$^{\pluscross}$ & 0.109$^{\pluscross}$ & 0.755$^{\pluscross}$ & 0.250$^{\pluscross}$ & 0.205$^{\pluscross}$ &  \textbackslash & \textbackslash & \textbackslash \\ \hline
CUT+SCC \cite{guo2022alleviating} & 0.572 & 0.185 & 0.110 & 0.699 & 0.263 & 0.182 & 27.34 & 39.2 & 60.5 \\ \hline
SSC \cite{zhao2023unsupervised} & 0.654 & 0.186 & 0.113 & 0.714 & 0.263 & 0.184 & 27.19 & 41.8 & 62.1 \\ \hline
\rowcolor[gray]{.9} Ours & \bftab 0.693 & \bftab 0.205 & \bftab 0.135 & \bftab 0.790 & \bftab 0.266 & \bftab 0.213 & 27.15 & \bftab 45.7 & 63.7 \\ \hline
\end{tabular} } 
\end{table*}

\section{Experiments}

To demonstrate the effectiveness of the proposed SemST method, we
conduct a series of experiments involving image translation on various
datasets.  These experiments prove that our method can improve
performance by mitigating semantic distortions. Furthermore, we perform
testing to confirm that the refined synthetic images can effectively aid
the downstream semantic segmentation task and potentially serve as a
beneficial pre-training procedure for UDA. 

\subsection{Experimental Settings}

Our implementation is based on the source code of CUT
\cite{park2020contrastive}. We substitute the original loss
with the TS loss and the hDCE loss proposed in this work. Moreover, we
restructure the original network to become a multi-scale architecture.
The global crop parameter ($h_g, w_g$) and the local crop parameter ($h_l, w_g$)
are both set to 256 (see Figure \ref{fig:pipeline}). 

\begin{figure*} [ht]
\begin{center} \begin{tabular}{c}
\includegraphics[width=0.97 \linewidth, height=4.6cm]{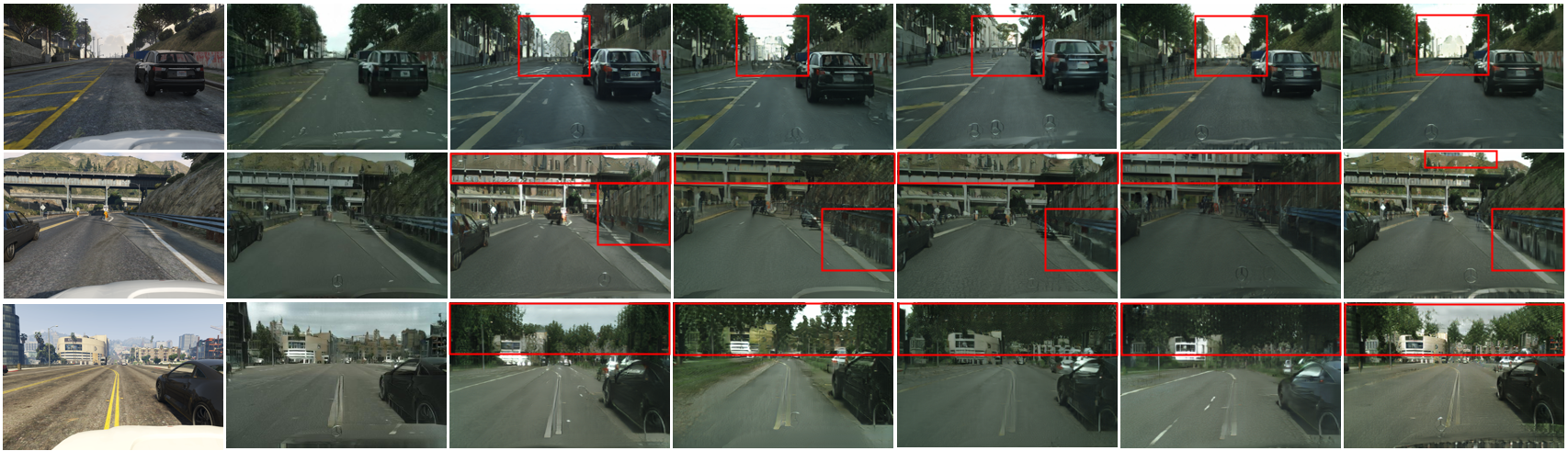} \\
\hspace{0.2cm} Input \hspace{1.5cm} \textbf{Ours} \hspace{1.5cm} CUT 
\hspace{1.6cm} SCC \hspace{1.6cm} SRC \hspace{1.7cm} SSC \hspace{1.1cm} 
CycleGAN \\ 
[\smallskipamount]
\includegraphics[width=0.97 \linewidth, height=2.9cm]{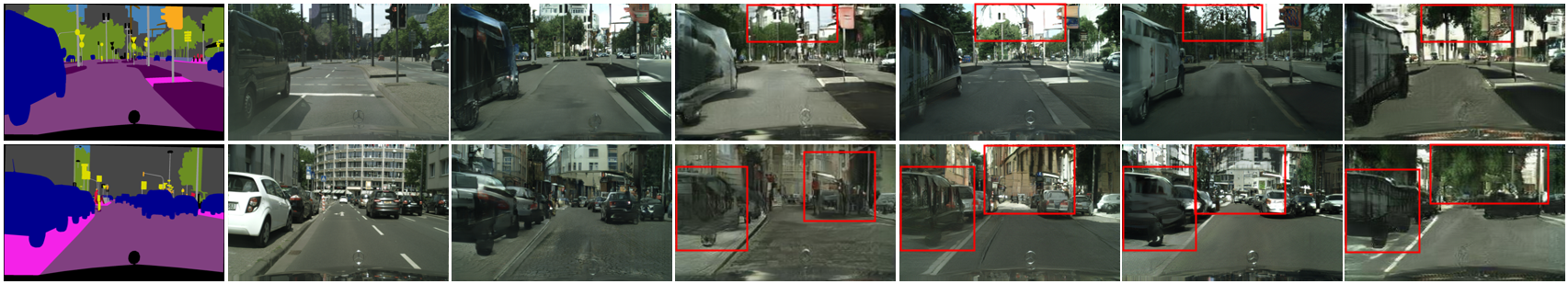} \\
\hspace{0.2cm} Input \hspace{0.9cm} Gound Truth \hspace{1cm} \textbf{Ours} \hspace{1.5cm} CUT \hspace{1.6cm} SRC \hspace{1.1cm} NEGCUT \hspace{0.8cm} CycleGAN 
\end{tabular} \end{center}
\caption[1]{Qualitative visual comparison of images refined by our SemST
method and other benchmarking methods on GTA5 $\rightarrow$ Cityscapes (top)
and Parsing $\rightarrow$ Image (bottom). Our method reduces the semantic
distortion and has fewer artifacts highlighted by bounding boxes.}
\label{fig:gta2cs}
\end{figure*}

\begin{figure*} [ht]
\begin{center} \begin{tabular}{c}
\includegraphics[height = 7.3cm, width=0.87 \linewidth]{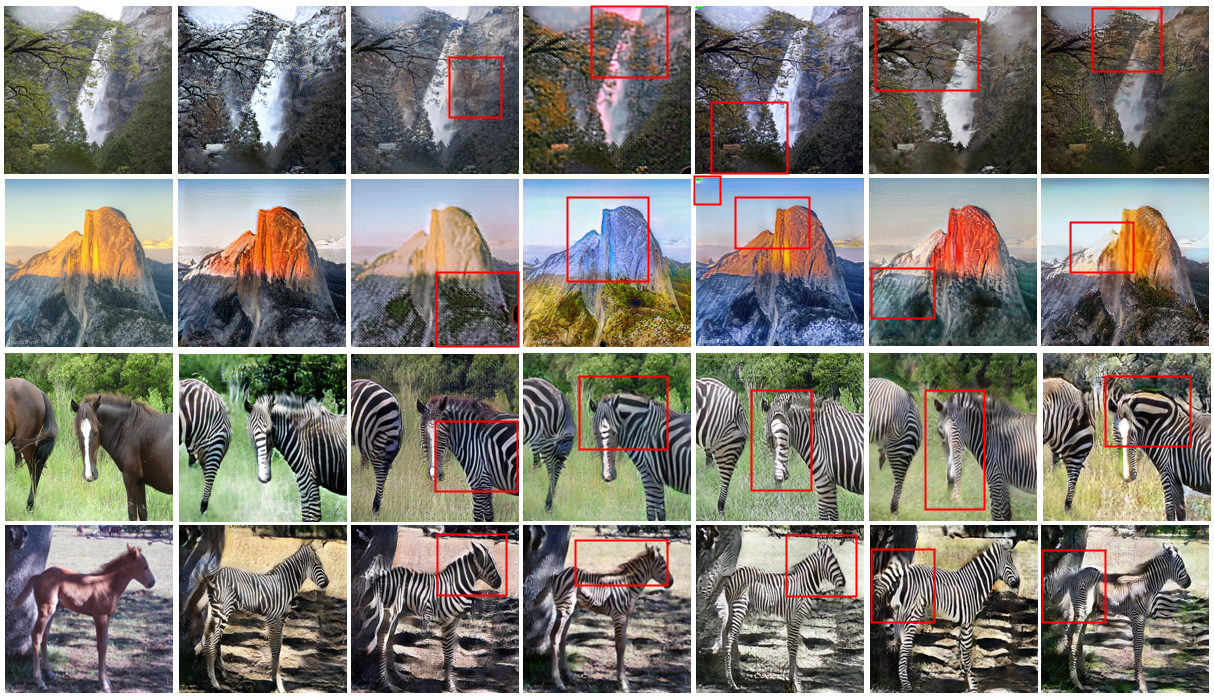} \\
\hspace{0.3cm} Input \hspace{1.3cm} \textbf{Ours} \hspace{1cm} NEGCUT \hspace{0.5cm} CycleGAN \hspace{0.8cm} SRC \hspace{1.2cm} CUT \hspace{1.1cm} SRUNIT 
\end{tabular} \end{center}
\caption[1]{Qualitative visual comparison of images refined by our SemST
method versus other benchmarking methods on summer $\rightarrow$ winter and
horse $\rightarrow$ zebra.  In the former, our results realistically cloaked leaves and mountains with snow, exhibiting superior or comparable authentic color representations. In the latter, we generate better or comparable natural color tones and preserve the horse's morphology. Generally, our outcomes contain fewer artifacts. }\label{fig:horse2zebra}
\end{figure*}

\subsection{Image-to-Image Translation}

To validate the effectiveness of our SemST method in the image-to-image
translation task and its capability to maintain semantic consistency
between input and output images, we have extensively tested it on
multiple datasets, including paired datasets (e.g., photo to map) and
unpaired datasets (GTA to Cityscapes, etc.). Both quantitative results
(see Table \ref{tab:img_trans}) and qualitative results (see Figure
\ref{fig:gta2cs} and \ref{fig:horse2zebra}) demonstrate its superior performance.  These results
are elaborated below. 

\subsubsection{Simulation to Real: GTA5 $\rightarrow$ Cityscapes} 

To prove our model can enhance the realism of synthetic images by
converting them into the domain of real-world captured images, we
convert the images from GTA5 \cite{richter2016playing} to Cityscapes
\cite{cordts2016cityscapes} domains. We train the model based on GTA5's
official training split, comprising 6,202 images with a resolution of
$1920 \times 1080$. In inference, we refine the first $500$ images in
the official test split and evaluate the performance by feeding them to
FCN-8s \cite{long2015fully} pre-trained on Cityscapes by pix2pix
\cite{isola2017image} to predict semantic label maps. We compute pixel
accuracy, class accuracy, and mean IoU by comparing the predicted and
ground-truth label maps. Higher scores indicate a similar distribution
between output and target images and consistent semantics between input
and output images. Thus, such a model can offer refined synthetic images
of higher quality and potentially benefit downstream semantic
segmentation. 

SemST significantly outperforms other benchmarking methods in all
metrics, achieving state-of-the-art performance, as shown in Table
\ref{tab:img_trans}. Exemplary images translated by different methods
are visualized in Figure \ref{fig:gta2cs}.  Evidently, SemST attains
superior visual quality by preserving the texture and structure of
synthetic input images and, consequently, maintaining the semantic
information. In contrast, semantic distortions exist in other
benchmarking methods.  For example, some sky region is converted to
vegetation or buildings, which are marked by red bounding boxes. 

\subsubsection{Parsing $\rightarrow$ Image} 

We train our model on the official training split of 3,975 and test on
the validation split of 500 images from Cityscapes, which has a
resolution of $2048 \times 1024$. Specifically, we transform semantic
label maps into corresponding images. Similar to the simulation-to-real
experiment, we assess different methods using metrics computed on
pre-trained FCN-8s \cite{isola2017image, long2015fully}.  Higher
evaluation metrics represent less semantic distortion between input and output. 

As shown in Table \ref{tab:img_trans}, SemST gives new state-of-the-art
performance in all metrics. Figure \ref{fig:gta2cs} provides a
qualitative comparison with other methods, demonstrating the capability
of SemST to produce finer borders and preserve each segmentation
region's semantics effectively. 

\subsubsection{Photo $\rightarrow$ Maps} 

We use the Maps dataset \cite{isola2017image} to further demonstrate the
performance of SemST on the I2I translation task. The dataset contains 2,194
pairs of aerial photo-to-map images, with 1,096 training and 1,098
testing pairs.  Following \cite{fu2019geometry}, we use RMSE and pixel
accuracy with threshold $\delta$ ($\delta_1 = 5$ and $\delta_2 = 10$) to
evaluate performance, where given ground truth pixel $p_i = (r_i, g_i,
b_i)$ and the prediction $\hat{p}_i = (\hat{r}_i, \hat{g}_i,
\hat{b}_i)$, pixel accuracy is computed by the indicator function
$\sum_{i=1}^{N} \mathbf{I} \{ max(|r_i-\hat{r}_i|, |g_i-\hat{g}_i|,
|b_i-\hat{b}_i|) < \delta \}$.

Again, Table \ref{tab:img_trans} shows that SemST has the best
performance regarding pixel accuracy with $\delta_1$.  While CycleGAN produces
lower RMSE and pixel accuracy with $\delta_2$ due to its cycle consistency loss, which is particularly
beneficial for the paired Maps dataset, SemST still generates the best
results among all contrastive-learning-based approaches. This
demonstrates the effectiveness of the proposed TS loss function and the
multi-scale framework in improving the image translation task,
highlighting the robustness and versatility of SemST. 

\subsubsection{Qualitative Results on Low-resolution Images} 

We provide more visual results on two popular datasets in Figure
\ref{fig:horse2zebra}.  They are the Horse $\rightarrow$ Zebra dataset
and the Summer $\rightarrow$ Winter dataset.  The former has unpaired
$1,067$ horse images and $1,334$ zebra images.  The latter contains
$1,231$ summer scenes and $962$ winter scenes in Yosemite.  There exists
a difference in semantic statistics between the two domains for each
dataset.  Since the resolution of images is small (i.e., $256 \times
256$), we use the single-scale method to predict results directly.
These experiments demonstrate the effectiveness of our proposed loss
function in preserving semantics and the resulting images have better or
comparable quality compared to others. 


\begin{table*}[htbp]
\caption{The IoU performance comparison of UDA methods and their
enhanced variants by incorporating synthetic images refined by SemST
(highlighted in gray shadow) in training. All methods are based on
DeepLab-V2 with ResNet-101. Training with
images refined by SemST improves UDA, demonstrating the effectiveness of our
method in UDA by reducing image-level domain gap.} \label{tab:da} 
\centering
\scalebox{0.72} {
\begin{tabular}{c|ccccccccccccccccccc|c}\hline
Method & road & side. & buil. & wall & fence & pole & light & sign & veg. & terr. & sky & pers. & rider & car & truck & bus & train & mbike & bike & mIoU \\ \hline
\multicolumn{21}{c}{GTA5 $\rightarrow$ Cityscapes} \\ \hline
SePiCo \cite{xie2023sepico} & 95.6 & 69.2 & 89.0 & 40.8 & 38.6 & 44.3 & 56.3 & 64.4 & 88.3 & 46.5 & 88.6 & 73.1 & 47.6 & 90.7 & 58.9 & 53.8 & 5.4 & 22.4 & 43.8 & 58.8 \\ \hline 
\rowcolor[gray]{.9} +SemST & 95.8 & 70.2 & 88.4 & 45.9 & 37.2 & 45.6 & 53.4 & 62.1 & 86.9 & 39.9 & 82.3 & 70.9 & 47.0 & 90.5 & 54.5 & 60.4 & 0.1 & 48.4 & 62.2 & \textbf{60.1} \\ \hline
ProDA \cite{zhang2021prototypical} & 87.8 & 56.0 & 79.7 & 46.3 & 44.8 & 45.6 & 53.5 & 53.5 & 88.6 & 45.2 & 82.1 & 70.7 & 39.2 & 88.8 & 45.5 & 59.4 & 1.0 & 48.9 & 56.4 & 57.5 \\ \hline
\rowcolor[gray]{.9} +SemST & 91.8 & 62.6 & 83.6 & 43.5 & 45.5 & 47.7 & 54.2 & 56.4 & 88.7 & 49.2 & 82.6 & 70.5 & 38.6 & 88.9 & 47.1 & 56.4 & 0.1 & 47.7 & 56.5 & \textbf{58.5} \\ \hline
BDL \cite{li2019bidirectional} & 91.0 & 44.7 & 84.2 & 34.6 & 27.6 & 30.2 & 36.0 & 36.0 & 85.0 & 43.6 & 83.0 & 58.6 & 31.6 & 83.3 & 35.3 & 49.7 & 3.3 & 28.8 & 35.6 & 48.5 \\ \hline
\rowcolor[gray]{.9} +SemST & 92.6 & 49.0 & 85.7 & 36.4 & 30.0 & 32.6 & 34.4 & 32.7 & 84.3 & 46.2 & 84.3 & 57.5 & 34.9 & 82.8 & 42.6 & 50.7 & 0.3 & 36.6 & 39.5 & \textbf{50.2}  \\ \hline

\multicolumn{21}{c}{SYNTHIA $\rightarrow$ Cityscapes} \\ \hline
SePiCo \cite{xie2023sepico} & 79.2 & 42.9 & 85.6 & 9.9 & 4.2 & 38.0 & 52.5 & 53.3 & 80.6 & - & 81.2 & 73.7 & 47.4 & 86.2 & - & 63.1 & - & 48.0 & 63.2 & 57.3 \\ \hline 
\rowcolor[gray]{.9} +SemST & 79.5 & 45.3 & 80.0 & 3.2 & 1.2 & 38.3 & 61.2 & 54.1 & 83.4 & - & 81.3 & 74.8 & 49.9 & 90.3 & - & 64.3 & - & 50.9 & 69.6 & \textbf{58.0} \\ \hline
ProDA \cite{zhang2021prototypical} & 87.8 & 45.7 & 84.6 & 37.1 & 0.6 & 44.0 & 54.6 & 37.0 & 88.1 & - & 84.4 & 74.2 & 24.3 & 88.2 & - & 51.1 & - & 40.5 & 45.6 & 55.5 \\ \hline 
\rowcolor[gray]{.9} +SemST & 86.5 & 43.2 & 90.3 & 37.2 & 0.1 & 46.1 & 53.5 & 36.2 & 92.9 & - & 87.9 & 80.1 & 29.1 & 86.1 & - & 56.8 & - & 41.1 & 48.2 & \textbf{57.2} \\ \hline

\end{tabular} } \end{table*}

\begin{figure*} [ht]
\begin{center} \begin{tabular}{c}
\includegraphics[width=0.97 \linewidth, height=4.5cm]{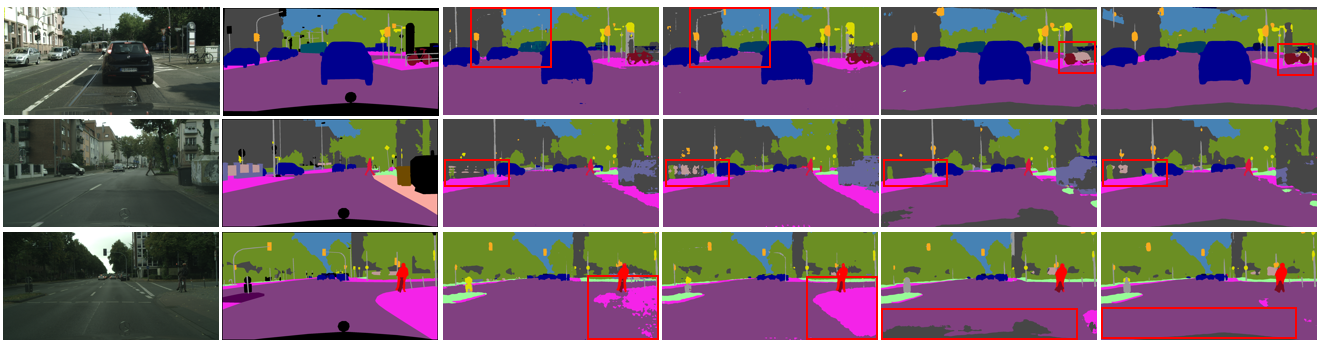} \\
\hspace{0.8cm} Image \hspace{1.6cm} Label \hspace{1.5cm} SePiCo \hspace{1.1cm} 
\textbf{SePiCo+SemST} \hspace{1.1cm} ProDA \hspace{1cm} \textbf{ProDA+SemST}
\end{tabular} \end{center}
\caption[1]{Qualitative visual comparison of results from domain
adaptation on GTA5 $\rightarrow$ Cityscapes using benchmarking methods
and those methods trained in combination with SemST-refined images.  The
latter ones have more accurate label predictions and refined borders, as
highlighted within the red bounding boxes.} \label{fig:da}
\end{figure*}

\subsection{Enhancing Semantic Segmentation} 

As discussed in Section \ref{sec:ref}, training on refined synthetic
images can enhance the downstream semantic segmentation task on
real-world datasets. Here, we demonstrate images refined by SemST can
assist unsupervised domain adaptation (UDA) by incorporating them in the training of domain
adaptation networks. Specifically, we train UDA models using images from the source domain and output images obtained by our
proposed domain mapper (i.e., refined synthetic images). We compare the
IoU scores obtained by different UDA methods and their enhanced variants
achieved by incorporating SemST-refined images into the training process
in Table \ref{tab:da}. The results are discussed below.

\subsubsection{GTA5 $\rightarrow$ Cityscapes}

We refine images from the GTA5 dataset to those in the Cityscapes
dataset and, then, include the refined images in the training of UDA
methods. Experimental results show that training with SemST-refined
synthetic images improves mIoU on different UDA methods, which indicates
that SemST can be potentially employed as a beneficial pre-training for
domain adaptation.  Another observation is the effect of class imbalance
on semantic segmentation performance.  Specifically, the failure in
predicting train class results from the low probability of them in class
distribution and different appearances across domains. In contrast,
success in road class prediction comes from high probability and similar
features across domains. 

\subsubsection{SYNTHIA $\rightarrow$ Cityscapes} 

We experiment on another source dataset called SYNTHIA-RAND-CITYSCAPES.
It is a subset of the synthetic urban scene dataset known as SYNTHIA
\cite{ros2016synthia}. It contains 9,400 images of resolution $1280
\times 760$ and 16 common semantic annotations with Cityscapes. After
refining the SYNTHIA dataset to the Cityscapes domain by SemST and
subsequently training the domain adaptor with refined images, we observe
a performance improvement. Experiments in Table \ref{tab:da}
showcase an improvement in the mIoU scores after training on refined
images. 

\begin{figure*} [ht]
\begin{center} \begin{tabular}{c}
\includegraphics[height=3cm, width=0.97 \linewidth]{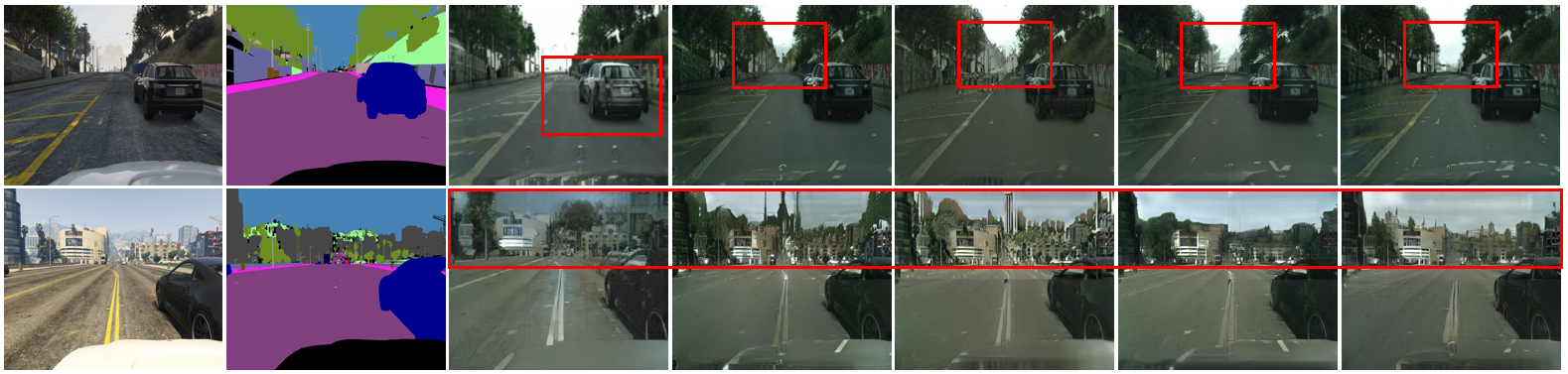} \\
\hspace{0.3cm} Input \hspace{1.2cm} Label \hspace{0.7cm} w/o Multiscale \hspace{0.2cm} w/o hDCE Loss \hspace{0.4cm} $\lambda_{TS}=0$ \hspace{1cm} $\lambda_{TS}=1$ \hspace{0.8cm} $\mathbf{\lambda_{TS}=2}$
\end{tabular} \end{center}
\caption[1]{Qualitative ablation study for GTA5 to Cityscapes. The red bounding boxes indicate artifacts. Removing multi-scale prediction yields inaccurate predictions and blurry results. Eliminating either the hDCE or TS loss introduces more artifacts and semantic distortion. As the weight of TS loss increases, semantic distortion is reduced.}\label{fig:ablation}
\end{figure*}

\begin{table}[htb]
\caption{Quantitative ablation study demonstrating the contributions of different components
}\label{table:ablation}
\centering
\resizebox*{\linewidth}{!}{
\begin{tabular}{c||c||c||ccc}\hline
\multirow{2}{*}{Methods}& \multicolumn{2}{c||}{w/o Components} & \multicolumn{3}{c}{Weights of TS Loss} \\ \cline{2-6} 
& Multiscale & hDCE Loss & 0 & 1 & \textbf{2} \\ \hline
pixel acc $\uparrow$  & 0.645 & 0.624 & 0.598  & 0.679 & \textbf{0.693} \\ \hline
class acc $\uparrow$ & 0.182 & 0.175 & 0.169 & 0.198 & \textbf{0.205} \\ \hline 
mean IoU $\uparrow$ & 0.115 & 0.113 & 0.110 & 0.126 & \textbf{0.135} \\ \hline
\end{tabular}}
\end{table}

\section{Ablation Study}

We examine the contribution of each individual component by excluding
them and varying the hyperparameters of the TS loss, denoted as
$\lambda_{TS}$, in the context of the GTA5 to Cityscapes task.  The
results of our investigations are presented in Table
\ref{table:ablation} and Figure \ref{fig:ablation}. Notably, removing any
component results in decreased performance. Specifically, multi-scale
prediction ensures superior performance on high-resolution images by local and global information learning. The hDCE loss alleviates the NPC effect and
enables more efficient learning from informative hard negative samples.
Furthermore, increasing the value of $\lambda_{TS}$ enhances performance
by mitigating semantic distortion. However, caution is needed in
selecting the magnitude of $\lambda_{TS}$, as excessively high values
would prompt the model to prioritize input-output consistency at the
potential expense of neglecting the style information learned from the
target domain. 

\section{Conclusion}\label{sec:Conclusions}

A multi-scale image translation method that preserves the semantic
consistency between input and output images, called SemST, was presented
in this work.  The multi-scale framework was used to predict local
detail and global context, which improves performance and enables the
application to higher-resolution images for UDA.  Semantic consistency
was achieved by introducing TS loss that aligns semantics between input
and output images by maximizing their mutual information in a shared
embedding space.  Extensive experiments were conducted to demonstrate
the state-of-the-art performance of SemST in image translation and its
value in facilitating UDA was also validated. 



{
\bibliographystyle{ieee_fullname}
\bibliography{egbib}

\begin{thebibliography}{10}\itemsep=-1pt

\bibitem{amodio2019travelgan}
Matthew Amodio and Smita Krishnaswamy.
\newblock Travelgan: Image-to-image translation by transformation vector
  learning.
\newblock In {\em Proceedings of the ieee/cvf conference on computer vision and
  pattern recognition}, pages 8983--8992, 2019.

\bibitem{atapattu2019improving}
Charith Atapattu and Banafsheh Rekabdar.
\newblock Improving the realism of synthetic images through a combination of
  adversarial and perceptual losses.
\newblock In {\em 2019 International Joint Conference on Neural Networks
  (IJCNN)}, pages 1--7. IEEE, 2019.

\bibitem{bachman2019learning}
Philip Bachman, R~Devon Hjelm, and William Buchwalter.
\newblock Learning representations by maximizing mutual information across
  views.
\newblock {\em Advances in neural information processing systems}, 32, 2019.

\bibitem{benaim2017one}
Sagie Benaim and Lior Wolf.
\newblock One-sided unsupervised domain mapping.
\newblock {\em Advances in neural information processing systems}, 30, 2017.

\bibitem{chen2021simpler}
Junya Chen, Zhe Gan, Xuan Li, Qing Guo, Liqun Chen, Shuyang Gao, Tagyoung
  Chung, Yi Xu, Belinda Zeng, Wenlian Lu, et~al.
\newblock Simpler, faster, stronger: Breaking the log-k curse on contrastive
  learners with flatnce.
\newblock {\em arXiv preprint arXiv:2107.01152}, 2021.

\bibitem{chen2017deeplab}
Liang-Chieh Chen, George Papandreou, Iasonas Kokkinos, Kevin Murphy, and Alan~L
  Yuille.
\newblock Deeplab: Semantic image segmentation with deep convolutional nets,
  atrous convolution, and fully connected crfs.
\newblock {\em IEEE transactions on pattern analysis and machine intelligence},
  40(4):834--848, 2017.

\bibitem{chen2016attention}
Liang-Chieh Chen, Yi Yang, Jiang Wang, Wei Xu, and Alan~L Yuille.
\newblock Attention to scale: Scale-aware semantic image segmentation.
\newblock In {\em Proceedings of the IEEE conference on computer vision and
  pattern recognition}, pages 3640--3649, 2016.

\bibitem{chen2020simple}
Ting Chen, Simon Kornblith, Mohammad Norouzi, and Geoffrey Hinton.
\newblock A simple framework for contrastive learning of visual
  representations.
\newblock In {\em International conference on machine learning}, pages
  1597--1607. PMLR, 2020.

\bibitem{cordts2016cityscapes}
Marius Cordts, Mohamed Omran, Sebastian Ramos, Timo Rehfeld, Markus Enzweiler,
  Rodrigo Benenson, Uwe Franke, Stefan Roth, and Bernt Schiele.
\newblock The cityscapes dataset for semantic urban scene understanding.
\newblock In {\em Proceedings of the IEEE conference on computer vision and
  pattern recognition}, pages 3213--3223, 2016.

\bibitem{fu2019geometry}
Huan Fu, Mingming Gong, Chaohui Wang, Kayhan Batmanghelich, Kun Zhang, and
  Dacheng Tao.
\newblock Geometry-consistent generative adversarial networks for one-sided
  unsupervised domain mapping.
\newblock In {\em Proceedings of the IEEE/CVF Conference on Computer Vision and
  Pattern Recognition}, pages 2427--2436, 2019.

\bibitem{guo2022alleviating}
Jiaxian Guo, Jiachen Li, Huan Fu, Mingming Gong, Kun Zhang, and Dacheng Tao.
\newblock Alleviating semantics distortion in unsupervised low-level
  image-to-image translation via structure consistency constraint.
\newblock In {\em Proceedings of the IEEE/CVF Conference on Computer Vision and
  Pattern Recognition}, pages 18249--18259, 2022.

\bibitem{he2020momentum}
Kaiming He, Haoqi Fan, Yuxin Wu, Saining Xie, and Ross Girshick.
\newblock Momentum contrast for unsupervised visual representation learning.
\newblock In {\em Proceedings of the IEEE/CVF conference on computer vision and
  pattern recognition}, pages 9729--9738, 2020.

\bibitem{hoffman2018cycada}
Judy Hoffman, Eric Tzeng, Taesung Park, Jun-Yan Zhu, Phillip Isola, Kate
  Saenko, Alexei Efros, and Trevor Darrell.
\newblock Cycada: Cycle-consistent adversarial domain adaptation.
\newblock In {\em International conference on machine learning}, pages
  1989--1998. Pmlr, 2018.

\bibitem{hoyer2022hrda}
Lukas Hoyer, Dengxin Dai, and Luc Van~Gool.
\newblock Hrda: Context-aware high-resolution domain-adaptive semantic
  segmentation.
\newblock In {\em Computer Vision--ECCV 2022: 17th European Conference, Tel
  Aviv, Israel, October 23--27, 2022, Proceedings, Part XXX}, pages 372--391.
  Springer, 2022.

\bibitem{hu2021adco}
Qianjiang Hu, Xiao Wang, Wei Hu, and Guo-Jun Qi.
\newblock Adco: Adversarial contrast for efficient learning of unsupervised
  representations from self-trained negative adversaries.
\newblock In {\em Proceedings of the IEEE/CVF Conference on Computer Vision and
  Pattern Recognition}, pages 1074--1083, 2021.

\bibitem{isola2017image}
Phillip Isola, Jun-Yan Zhu, Tinghui Zhou, and Alexei~A Efros.
\newblock Image-to-image translation with conditional adversarial networks.
\newblock In {\em Proceedings of the IEEE conference on computer vision and
  pattern recognition}, pages 1125--1134, 2017.

\bibitem{jia2021semantically}
Zhiwei Jia, Bodi Yuan, Kangkang Wang, Hong Wu, David Clifford, Zhiqiang Yuan,
  and Hao Su.
\newblock Semantically robust unpaired image translation for data with
  unmatched semantics statistics.
\newblock In {\em Proceedings of the IEEE/CVF International Conference on
  Computer Vision}, pages 14273--14283, 2021.

\bibitem{jung2022exploring}
Chanyong Jung, Gihyun Kwon, and Jong~Chul Ye.
\newblock Exploring patch-wise semantic relation for contrastive learning in
  image-to-image translation tasks.
\newblock In {\em Proceedings of the IEEE/CVF Conference on Computer Vision and
  Pattern Recognition}, pages 18260--18269, 2022.

\bibitem{kim2017learning}
Taeksoo Kim, Moonsu Cha, Hyunsoo Kim, Jung~Kwon Lee, and Jiwon Kim.
\newblock Learning to discover cross-domain relations with generative
  adversarial networks.
\newblock In {\em International conference on machine learning}, pages
  1857--1865. PMLR, 2017.

\bibitem{lee2018diverse}
Hsin-Ying Lee, Hung-Yu Tseng, Jia-Bin Huang, Maneesh Singh, and Ming-Hsuan
  Yang.
\newblock Diverse image-to-image translation via disentangled representations.
\newblock In {\em Proceedings of the European conference on computer vision
  (ECCV)}, pages 35--51, 2018.

\bibitem{lee2020drit++}
Hsin-Ying Lee, Hung-Yu Tseng, Qi Mao, Jia-Bin Huang, Yu-Ding Lu, Maneesh Singh,
  and Ming-Hsuan Yang.
\newblock Drit++: Diverse image-to-image translation via disentangled
  representations.
\newblock {\em International Journal of Computer Vision}, 128:2402--2417, 2020.

\bibitem{li2020prototypical}
Junnan Li, Pan Zhou, Caiming Xiong, and Steven~CH Hoi.
\newblock Prototypical contrastive learning of unsupervised representations.
\newblock {\em arXiv preprint arXiv:2005.04966}, 2020.

\bibitem{li2019bidirectional}
Yunsheng Li, Lu Yuan, and Nuno Vasconcelos.
\newblock Bidirectional learning for domain adaptation of semantic
  segmentation.
\newblock In {\em Proceedings of the IEEE/CVF Conference on Computer Vision and
  Pattern Recognition}, pages 6936--6945, 2019.

\bibitem{long2015fully}
Jonathan Long, Evan Shelhamer, and Trevor Darrell.
\newblock Fully convolutional networks for semantic segmentation.
\newblock In {\em Proceedings of the IEEE conference on computer vision and
  pattern recognition}, pages 3431--3440, 2015.

\bibitem{ma2021coarse}
Haoyu Ma, Xiangru Lin, Zifeng Wu, and Yizhou Yu.
\newblock Coarse-to-fine domain adaptive semantic segmentation with photometric
  alignment and category-center regularization.
\newblock In {\em Proceedings of the IEEE/CVF Conference on Computer Vision and
  Pattern Recognition}, pages 4051--4060, 2021.

\bibitem{ma2022i2f}
Haoyu Ma, Xiangru Lin, and Yizhou Yu.
\newblock I2f: A unified image-to-feature approach for domain adaptive semantic
  segmentation.
\newblock {\em IEEE Transactions on Pattern Analysis and Machine Intelligence},
  2022.

\bibitem{musto2020semantically}
Luigi Musto and Andrea Zinelli.
\newblock Semantically adaptive image-to-image translation for domain
  adaptation of semantic segmentation.
\newblock {\em arXiv preprint arXiv:2009.01166}, 2020.

\bibitem{oord2018representation}
Aaron van~den Oord, Yazhe Li, and Oriol Vinyals.
\newblock Representation learning with contrastive predictive coding.
\newblock {\em arXiv preprint arXiv:1807.03748}, 2018.

\bibitem{park2020contrastive}
Taesung Park, Alexei~A Efros, Richard Zhang, and Jun-Yan Zhu.
\newblock Contrastive learning for unpaired image-to-image translation.
\newblock In {\em Computer Vision--ECCV 2020: 16th European Conference,
  Glasgow, UK, August 23--28, 2020, Proceedings, Part IX 16}, pages 319--345.
  Springer, 2020.

\bibitem{richter2016playing}
Stephan~R Richter, Vibhav Vineet, Stefan Roth, and Vladlen Koltun.
\newblock Playing for data: Ground truth from computer games.
\newblock In {\em Computer Vision--ECCV 2016: 14th European Conference,
  Amsterdam, The Netherlands, October 11-14, 2016, Proceedings, Part II 14},
  pages 102--118. Springer, 2016.

\bibitem{robinson2020contrastive}
Joshua Robinson, Ching-Yao Chuang, Suvrit Sra, and Stefanie Jegelka.
\newblock Contrastive learning with hard negative samples.
\newblock {\em arXiv preprint arXiv:2010.04592}, 2020.

\bibitem{ros2016synthia}
German Ros, Laura Sellart, Joanna Materzynska, David Vazquez, and Antonio~M
  Lopez.
\newblock The synthia dataset: A large collection of synthetic images for
  semantic segmentation of urban scenes.
\newblock In {\em Proceedings of the IEEE conference on computer vision and
  pattern recognition}, pages 3234--3243, 2016.

\bibitem{shen2023study}
Tingwei Shen, Ganning Zhao, and Suya You.
\newblock A study on improving realism of synthetic data for machine learning,
  2023.

\bibitem{shrivastava2017learning}
Ashish Shrivastava, Tomas Pfister, Oncel Tuzel, Joshua Susskind, Wenda Wang,
  and Russell Webb.
\newblock Learning from simulated and unsupervised images through adversarial
  training.
\newblock In {\em Proceedings of the IEEE conference on computer vision and
  pattern recognition}, pages 2107--2116, 2017.

\bibitem{theiss2022unpaired}
Justin Theiss, Jay Leverett, Daeil Kim, and Aayush Prakash.
\newblock Unpaired image translation via vector symbolic architectures.
\newblock In {\em European Conference on Computer Vision}, pages 17--32.
  Springer, 2022.

\bibitem{wang2021instance}
Weilun Wang, Wengang Zhou, Jianmin Bao, Dong Chen, and Houqiang Li.
\newblock Instance-wise hard negative example generation for contrastive
  learning in unpaired image-to-image translation.
\newblock In {\em Proceedings of the IEEE/CVF International Conference on
  Computer Vision}, pages 14020--14029, 2021.

\bibitem{wei2020co2}
Chen Wei, Huiyu Wang, Wei Shen, and Alan Yuille.
\newblock Co2: Consistent contrast for unsupervised visual representation
  learning.
\newblock {\em arXiv preprint arXiv:2010.02217}, 2020.

\bibitem{xie2023sepico}
Binhui Xie, Shuang Li, Mingjia Li, Chi~Harold Liu, Gao Huang, and Guoren Wang.
\newblock Sepico: Semantic-guided pixel contrast for domain adaptive semantic
  segmentation.
\newblock {\em IEEE Transactions on Pattern Analysis and Machine Intelligence},
  2023.

\bibitem{yamada2013relative}
Makoto Yamada, Taiji Suzuki, Takafumi Kanamori, Hirotaka Hachiya, and Masashi
  Sugiyama.
\newblock Relative density-ratio estimation for robust distribution comparison.
\newblock {\em Neural computation}, 25(5):1324--1370, 2013.

\bibitem{yang2022mutual}
Chuanguang Yang, Zhulin An, Linhang Cai, and Yongjun Xu.
\newblock Mutual contrastive learning for visual representation learning.
\newblock In {\em Proceedings of the AAAI Conference on Artificial
  Intelligence}, volume~36, pages 3045--3053, 2022.

\bibitem{yeh2022decoupled}
Chun-Hsiao Yeh, Cheng-Yao Hong, Yen-Chi Hsu, Tyng-Luh Liu, Yubei Chen, and Yann
  LeCun.
\newblock Decoupled contrastive learning.
\newblock In {\em Computer Vision--ECCV 2022: 17th European Conference, Tel
  Aviv, Israel, October 23--27, 2022, Proceedings, Part XXVI}, pages 668--684.
  Springer, 2022.

\bibitem{yi2017dualgan}
Zili Yi, Hao Zhang, Ping Tan, and Minglun Gong.
\newblock Dualgan: Unsupervised dual learning for image-to-image translation.
\newblock In {\em Proceedings of the IEEE international conference on computer
  vision}, pages 2849--2857, 2017.

\bibitem{zhang2021prototypical}
Pan Zhang, Bo Zhang, Ting Zhang, Dong Chen, Yong Wang, and Fang Wen.
\newblock Prototypical pseudo label denoising and target structure learning for
  domain adaptive semantic segmentation.
\newblock In {\em Proceedings of the IEEE/CVF conference on computer vision and
  pattern recognition}, pages 12414--12424, 2021.

\bibitem{zhang2019harmonic}
Rui Zhang, Tomas Pfister, and Jia Li.
\newblock Harmonic unpaired image-to-image translation.
\newblock {\em arXiv preprint arXiv:1902.09727}, 2019.

\bibitem{zhao2023unsupervised}
Ganning Zhao, Tingwei Shen, Suya You, and C-C~Jay Kuo.
\newblock Unsupervised synthetic image refinement via contrastive learning and
  consistent semantic and structure constraints.
\newblock {\em arXiv preprint arXiv:2304.12591}, 2023.

\bibitem{zhu2017unpaired}
Jun-Yan Zhu, Taesung Park, Phillip Isola, and Alexei~A Efros.
\newblock Unpaired image-to-image translation using cycle-consistent
  adversarial networks.
\newblock In {\em Proceedings of the IEEE international conference on computer
  vision}, pages 2223--2232, 2017.

\end{thebibliography}
}

\end{document}